\documentclass[11pt]{article}
\usepackage{nodalida2025}
\usepackage{times}
\usepackage{url}
\usepackage{latexsym}

\usepackage[T1]{fontenc}
\usepackage{graphicx}
\usepackage[utf8]{inputenc} % allow utf-8 input
\usepackage[T1]{fontenc}    % use 8-bit T1 fonts
\usepackage{hyperref}       % hyperlinks
\usepackage{url}            % simple URL typesetting
\usepackage{booktabs}       % professional-quality tables
\usepackage{amsfonts}       % blackboard math symbols
\usepackage{nicefrac}       % compact symbols for 1/2, etc.
\usepackage{microtype}      % microtypography
\usepackage{xcolor}         % colors
\usepackage{amsmath} 
\usepackage{multirow}
\usepackage{booktabs,arydshln}
\usepackage{subcaption}
\usepackage[colorinlistoftodos,bordercolor=orange,backgroundcolor=orange!20,linecolor=orange,textsize=scriptsize]{todonotes}
\makeatletter
\def\adl@drawiv#1#2#3{%
        \hskip.5\tabcolsep
        \xleaders#3{#2.5\@tempdimb #1{1}#2.5\@tempdimb}%
                #2\z@ plus1fil minus1fil\relax}
\newcommand{\cdashlinel}[1]{%
  \noalign{\vskip\aboverulesep
           \global\let\@dashdrawstore\adl@draw
           \global\let\adl@draw\adl@drawiv}
  \cdashline{#1}
  \noalign{\global\let\adl@draw\@dashdrawstore
           \vskip\belowrulesep}}
\makeatother

\usepackage{xspace}

\begin{document}
\title{\LARGE When are 1.58 bits enough? A Bottom-up Exploration of BitNet~Quantization}

% Nielsen
% Galke
% Schneider-Kamp
% \author{Jacob Nielsen \\
%   Affiliation / Address line 1 \\
%   Affiliation / Address line 2 \\
%   Affiliation / Address line 3 \\
%   {\tt jacn@imada.sdu.dk} \\\And
%   Lukas Galke \\
%   Affiliation / Address line 1 \\
%   Affiliation / Address line 2 \\
%   Affiliation / Address line 3 \\
%   {\tt galke@imada.sdu.dk} \\\And
%   Peter Schneider-Kamp \\
%   Affiliation / Address line 1 \\
%   Affiliation / Address line 2 \\
%   Affiliation / Address line 3 \\
%   {\tt petersk@imada.sdu.dk} \\}

\author{\Large Jacob Nielsen \qquad Lukas Galke \qquad Peter Schneider-Kamp \\ 
Department of Mathematics and Computer Science, University of Southern Denmark, Odense, Denmark \\ 
\texttt{\{jacn,galke,petersk\}@imada.sdu.dk}}

\date{}

% PRE NODALIDA Stuff below
% %
% \titlerunning{Exploring BitNet b1.58}
% % If the paper title is too long for the running head, you can set
% % an abbreviated paper title here
% %
% \author{Author 0 \inst{1}\orcidID{0000-0000-0000-0000} \and\\
% Author 1 \inst{1}\orcidID{0000-0000-0000-0000}}
% %
% \authorrunning{Author 0 \& Author 1}
% \institute{Department of Mathematics and Computer Science, University of Southern Denmark, Odense, Denmark\\
% \email{\{Author0 , Author1\}@imada.sdu.dk}}
% %
\maketitle              % typeset the header of the contribution
\begin{abstract}
% The abstract should briefly summarize the contents of the paper in
Contemporary machine learning models, such as language models, are powerful, but come with immense resource requirements both at training and inference time. It has been shown that decoder-only language models can be trained to a competitive state with ternary weights (1.58 bits per weight), facilitating efficient inference.  Here, we start our exploration with non-transformer model architectures, investigating 1.58-bit training for multi-layer perceptrons and graph neural networks. 
Then, we explore 1.58-bit training in other transformer-based language models, namely encoder-only and encoder-decoder models.
Our results show that in all of these settings, 1.58-bit training is on par with or sometimes even better than the standard 32/16-bit models. 
\end{abstract}
%
%
% \keywords{deep learning \and quantization-aware training \and green machine learning \and small language models \and image classification.}
%

% \todo{related work: https://arxiv.org/pdf/2405.02543, https://arxiv.org/abs/2408.13402 }
% \todo{ started as a means of improving latency, throughput, memory consumption, and energy consumption of notoriously hungy LLMs. then extension to multi-model LLMs. This top-down has established that it works. Circumstantial evidence suggests it is not limited to large models, though. [on-par performance for small vision models, scaling law for small language models]. in this work we aim to explore bitnet 1.58 more systematically using a bottom-up approach, yielding insights into the capabilities of 1.58 bit qantization-aware training and its limitations.}

\section{Introduction}
Large Language Models (LLMs) have dominated the field of natural language processing in recent years. The growing capabilities and applications of these demonstrate potential for enriching our society. However, LLMs come with deployments obstacles, especially regarding memory use, latency, and throughput. What is more, considerations regarding LLM's environmental impact in terms of energy consumption are threatening their uptake.

% \jacob{Come back to the next two sections.}
Post-training quantisation methods have been proposed, including but not limited to, Generative Pre-trained Transformer Quantization \cite{frantar2023gptq} and Activation-aware Weight Quantization \cite{lin2024awq}. Post-training quantization has also been employed in other domain such as vision models \cite{li2023vit}. However, post-training quantization inherently comes at the cost of precision. 

An alternative to post-training quantization is quantization-aware training such as LLM-QAT \cite{liu2023llmqat} and QA-LoRA \cite{xu2023qaloraquantizationawarelowrankadaptation}. Here, as the training optimizes the quantized weights, there is no loss of precision when using the quantized model for inference. Recent works on 1-bit~\cite{wang2023bitnet} and 1.58-bit~\cite{ma2024era} quantization-aware training architectures have demonstrated the potential of training in a very low-bit representation while still maintaining most or all of the LLM's performance.

The 1.58-bit quantization-aware training architecture introduced in BitNet b1.58 \cite{ma2024era}, which quantizes the weights of linear layers to the values $-1$, $0$, or $1$ during forward passes, yields only minimal performance hits in 3B+ parameter LLMs. Building on this, recent work suggests that 1.58-bit training is promising for multi-modal architectures~\cite{sundaram2024llavaolmobitnet1b} % and encourage the community to find ways to do Post-Training Quantize/Quantization-Aware Finetune open weight pre-trained models to ternary domain
% From above 3B parameters, the 1.58-bit models trained from scratch perform just as well as 16-bit models. 
and spiking language models \cite{bal2024exploring}.

BitNet b1.58 Reloaded \cite{nielsen2024bitnetb158reloadedstateoftheart} investigates the b1.58 architecture in small vision and small language models, ranging from 100K to 48M parameters. The authors define a scaling law between 16-bit and 1.58-bit layers, showing that 1.58-bit can achieve similar performance to 16-bit training, even on these small lower-capacity networks. Furthermore, they show that employing the median (as a basis for quantization and rescaling) instead of the mean achieves even better performance in some settings, ostensibly by decreasing sensitivity to outliers.

However, the b1.58 architecture still poses a lot of unanswered questions, which require further investigation of its potential and limitations. Here,  we investigate 1.58-bit quantization-aware training in a bottom-up manner, starting from the classic exclusive or (X-OR) task and non-transformer architectures such as multi-layer perceptrons and graph neural networks. We further fill a gap in the literature by analyzing 1.58-bit quantization-aware training for encoder-only and encoder-decoder transformer-based language models, complementing prior work on decoder-only language models.

Our results establish that 1.58-bit weights can be employed as a drop-in replacement for Linear layers in a multitude of model architectures with no to minimal loss in performance. We find that in encoder-only language models, commensurate accuracy with 16-bit models can be obtained by increasing the hidden size throughout the model. Moreover, we demonstrate that using 1.58-bit quantization in the T5 architecture can outperform 16-bit models, hinting at a possible regularization effect.

In summary, our contributions are as follows:
\begin{itemize}
    \item A bottom-up exploration of 1.58-bit quantization-aware training, ranging from the X-OR problem (Section~\ref{sec:x-or}) and 1.58-bit multi-layer perceptrons for text classification (Section~\ref{sec:non-transformer:mlp}) to 1.58-bit graph neural networks for node classification (Section~\ref{sec:non-transformer:gnn}).
    \item Experiments on 1.58-bit encoder-only language models showing that an increase in model capacity can compensate for the lower bit precision -- yet sub-proportionally to the decrease in bit width (Section~\ref{sec:transformers:encoders}).
    \item Experiments on 1.58-bit encoder-decoder language models demonstrated superior performance on the 1.58-bit variant compared to their 16-bit counterpart, without the the need for parameter-compensating (Section~\ref{sec:transformer:endec}).
\end{itemize}

% \begin{itemize}
%     \item Activation range -> We don't see a big gain by increasing on Encoder-archs. 
%     \item Encoder performance 
%     \item Decoder Performance 
%     \item Encoder/Decoder
%     \begin{itemize}
%         \item Problem with $o$-projection in \texttt{SelfAttention.}
%         \item Investigate solution (probably later article)
%     \end{itemize}
% \end{itemize}

% In this work we investigate 1.58-bit quantization aware training for small language models (SLMs) and vision models ranging from 100K to 48M parameters. We introduce a variant of BitNet b1.58 that relies on the median rather than the mean of the absolute values of the weights. Through extensive experiments we investigate and compare the scaling, the learning-rate robustness, and the regularization properties of both 1.58-bit variants. Our work demonstrates that 1.58-bit quantization aware training can get close to state-of-the-art performance on SLMs and even exceed the state-of-the-art performance on vision models, opening a new avenue for research in this direction. This facilitates the deployment of SLMs and small vision models in low-ressource settings. Our implementation is available from GitHub\footnote{https://github.com/schneiderkamplab/bitlinear} and the Python Packacking Index\footnote{https://pypi.org/project/bitlinear/}.
% \input{sections/related_work}

\section{Background: b1.58 Quantization}\label{sec:method}
We recapitulate the basics of 1.58-bit quantization proposed by \citet{nielsen2024bitnetb158reloadedstateoftheart}, which generalizes the one from \cite{wang2023bitnet}.

The core of the 1.58-bit quantization scheme is to introduce a drop-in replacement for the \texttt{Linear} layers in common machine learning frameworks such as PyTorch, which we denote as \texttt{BitLinear}.
During training, the BitLinear layer retains 16-bit shadow weights. During the forward pass, the shadow weights are quantized to 1.58-bit precision which corresponds to ternary weights: $\{-1,0,1\}$. During the backward pass, the shadow weights are optimized via the straight-through estimator~\cite{DBLP:journals/corr/BengioLC13}.
Because forward passes are always conducted with quantized weights, we can drop the shadow weights when training concludes, using solely the ternary weights during inference.

The computation flow of a \texttt{BitLinear} layer follows a five-step procedure:
First, the activations are normalized via a parameter-free LayerNorm~\cite{ba2016layer}.
%\todo{We use layernorm instead of RMSNorm as it yielded better results in reloaded.}
%we already state CLEARLY that we use the Nielsen & Schneider-Kamp variant
We denote the layer normalization of input $I$ as $\hat{I}$. 

Second, the layer normalized activations are quantized to k-bit precision (usually $k=8$) via AbsMax quantization. We first calculate a scaling factor $x_\mathrm{scale}$ such that
$x_\mathrm{scale} = \frac{Q_b}{\max(|\hat{I}|) + \epsilon}$,
where $Q_b = 2^{k-1}$ is the range of the $k$ bits used for the quantized activations and $\epsilon$ is a small positive value preventing zero-division.
This ensures that all activations can be scaled to integer values $\{-Q_b, \ldots, Q_b-1\}$.
%\todo{I think we made a mistake in the DELTA paper - it should be -128 to 127, not -129 to 128 -- I guess that is because the original papers used $Q_b = 2^{k-1}-1$.}
We then employ AbsMax quantization on the activations as follows: 
$$\mathbf{x}_\mathrm{quant} =  \max(-Q_b, \min(Q_b-1, \operatorname{round}(\hat{\mathbf{I}} \cdot x_\mathrm{scale}))$$

\noindent
Third, the 16-bit shadow weights $W \in \mathcal{R}^{n\times m}$ are scaled and quantized to a ternary system of integer values in $\{-1, 0, 1\}$ (corresponding to 1.58 bits per weight
%\todo{1.58 refers to the entropy value when encoding the three values with equal prob. Is this the same?})
%Sure - guess we are fine without going there, though.
through a generic $\operatorname{AbsMeasure}$ quantization method~\cite{nielsen2024bitnetb158reloadedstateoftheart}. For this, we calculate a second scaling factor $w_\mathrm{scale} = \frac{1}{\operatorname{Measure}(|W|) + \epsilon}$, where $\operatorname{Measure}$ denotes either the mean or median function.
The quantized weights $\mathbf{W}_\mathrm{quant}$ can then be derived as:
$$
   \mathbf{W}_\mathrm{quant} = \max(-1, \min(1, \operatorname{round}(\mathbf{W} \cdot w_\mathrm{scale}))
$$

\noindent
Fourth, having quantized both the activations and the weights, we can conduct a forward pass with $\mathbf{x}_\mathrm{quant}$ and $\mathbf{W}_\mathrm{quant}$:
% \todo{good phrasing? Not optional to harvest the benfits. what about smth like: which using a specialized kernel can be optimizied significantly on bit-level, enhancing speed and use reduced model-size exploiting the properties of b158??}
%I gave it a go below.
$$
   \mathbf{y}_{\mathrm{quant}} = \mathbf{x}_{\mathrm{quant}} \cdot \mathbf{W}_{\mathrm{quant}} + \mathbf{b}
$$
where $\mathbf{b}$ is an optional bias term. To fully exploit the positive impacts on memory use, latency, and throughput, the forward pass ought to be carried out by a specialized kernel using bit-level operations.

We detach both $\mathbf{x}_\mathrm{quant}$ and $\mathbf{w}_\mathrm{quant}$ from the computation graph to facilitate a straight-through estimation of the gradients. The gradients will then be estimated with respect to the shadow weights, i.e., the 16-bit weights that were quantized via AbsMeasure in step 3.

Lastly, the result of this linear transformation is rescaled with the scaling factors $x_\mathrm{scale}$ and $w_\mathrm{scale}$ from steps 2 and 3, respectively.
Thus, to calculate the layer's final output $\mathbf{y}$, we rescale $y$ as follows:
$$
  \mathbf{y} = \frac{\mathbf{y}_{quant}}{w_{scale} \cdot x_{scale}}
$$

\newcommand\xor{X-OR\xspace}
\section{BitNet in Non-Transformer Models}\label{sec:non-transformer}
Prior work on BitNet quantization has predominantly focused on analyzing transformer models~\cite{wang2023bitnet,ma2024era,nielsen2024bitnetb158reloadedstateoftheart}. Here, we ask to what extent quantization-aware training is a feasible strategy for neural networks in general. We engage in a bottom-up exploration, starting from the popular \xor{} problem and a minimalist BitNet model, then advance to BitNet in multilayer perceptrons to carry out text classification from a bag-of-words representation, and, finally, explore BitNet for graph neural networks.

\subsection{Can 1.58-bit models solve X-OR?}\label{sec:x-or}
To better understand the dynamics of 1.58-bit training, we  explore the \xor{} problem, in which the model needs to learn the function of exclusive-or: Given two binary inputs, the task is to output 1 when exactly one of the inputs is 1 and 0 otherwise. The \xor{}-problem is particularly interesting because it is known that it cannot be solved by a purely linear model and requires at least one hidden layer with a non-linear activation.

In theory, ternary weights as in BitNet are sufficient to solve \xor{}. One possible solution would be to have a two hidden units, one assigning a positive weight to the first input and a negative weight to the second, with the second hidden unit vice-versa: $h_1 = \operatorname{ReLU}(x_1-x_2)$ and $h_2 = \operatorname{ReLU}(x_2-x_1)$. The output layer would then have two positive weights $y = h_1 + h_2$, solving the \xor{} problem.
However, whether these weights can be learned from data is a different question subject to experimentation.

\paragraph{Setup}
We set up the basic \xor{}-problem, while add two further noise inputs which do not affect the output and ought to be ignored (for instance, by assigning weight zero to those inputs). We are interested whether a BitNet model with a single hidden layer, i.e., two BitLinear layers with an intermediate ReLU activation, can learn a perfect solution to the \xor{} problem while ignoring the noise inputs. 
The weight range is set to 1.58 bits (-1, 0, or 1), the activation range is 8 bits. We train for 1k epochs on 5k examples with 4 features, two of which determine the target \xor output, the other two are noise. The learning rate is set to 0.01 unless noted otherwise. Optimization is carried out by Adam~\cite{kingma2014adam} against cross-entropy of two possible outputs.
%Evaluation is technicaly done on 5k separately sampled inputs -- yet all possible combinations of inputs have been seen at training.

% \todo{add:
% 2 hidden units MLP does find a solution
% 2 hidden units MLP-AbsMean does not find a solution with neither lr=0.001 nor lr=0.01

% 4 hidden units MLP-AbsMean does find a solution with 4 hidden units and lr=0.001
% 4 hidden units MLP-AbsMedian does not find a solution with 4 hidden units and lr=0.001 or lr=0.01
% }
\paragraph{Results} 
A standard MLP with one hidden layer of two hidden units solves X-OR. BitNet variants with two hidden units do not find a solution.

With 8 hidden units, and the mean quantization scheme, $\operatorname{AbsMean}$, the model finds a perfect solution (100\% accuracy) with exactly 4 nonzero parameters on the input layer and, in particular, all zero weights on the noise inputs.
However, with the median weight quantization ($\operatorname{AbsMedian}$), the model did not find a solution with 8 hidden units, converging at accuracy 86.8\%. The L1 norm of the \xor{} input weights
was 12 (out of 16) and the L1 norm of the noise input weights was 8 (out of 16).

When we increase the number of hidden units to 16, $\operatorname{AbsMedian}$
got perfect accuracy at the end of training, but the trajectory during training was unstable.
The L1 norm for \xor{} input weights was 24. The L1 norm for noise input weights
was 17. Output layer L1 norm 26 (out of 32).

With 32 hidden units, $\operatorname{AbsMedian}$ found a perfect solution (100\% accuracy) less fluctuation on the trajectory. The L1 norm of the weights for \xor{} inputs was 51 (out of 64). The L1 norm of the noise-input weights was 28 (out of 64). The L1 norm of the output layer
was 50 (out of 64).

Going back to 8 hidden units, but larger learning rate (0.1),
$\operatorname{AbsMedian}$ also finds a 100\% accurate solution. The  L1 norm of weights for
\xor{} inputs was 12 (out of 16), while the L1 norm of weights for noise
inputs: 5 (4 of them negative). The hidden-to-output layer had all non-zero
weights (L1 norm of 16). The bias parameters seemingly help to ignore parts of
the inputs in conjunction with ReLU activation. Note that, when the weights for
the noise inputs are negative, the bias term needs to compensate such that the 
noise inputs do not distort the sum with the \xor{} inputs.

In summary, both $\operatorname{AbsMean}$ and $\operatorname{AbsMedian}$ managed to solve the \xor{} problem. However, $\operatorname{AbsMedian}$ needed either a larger learning rate or a higher
amount of hidden units. 

\subsection{BitNet in Multilayer Perceptrons}\label{sec:non-transformer:mlp}
%\todo[color=red!20]{Lukas stopped here}
Next, we investigate whether a BitNet multilayer perceptron (MLP) can successfully learn to categorize texts based on bag-of-words features. 

\paragraph{Setup}
We use the same setup as a recent work on text classification \cite{galke-scherp-2022-bag}, where a  wide multilayer perceptron (WideMLP) on a bag-of-words representation had shown good results.
We train two BitNet variants (WideMLP-b1.58-mean and WideMLP-b1.58-median) of this WideMLP baseline.
The first layer of the WideMLP model is implemented as an embedding layer to avoid large matrix multiplications with the dimensions of the vocabulary size. Therefore, we quantize only the hidden-to-output layer of the two-layer architecture. Nevertheless, this leads to an interesting question how quantized fully-connected layers cope with non-quantized embedding layers.
We use 5 standard benchmark datasets: 20ng, R8, R52, ohsumed, and MR -- with four of them being topic classification and MR being sentiment classification.
We use the same hyperparameters as the baseline: training for 100 epoch with batch size 16.
  
\begin{table*}[ht]
    \centering
    \begin{tabular}{lcccccr}
    \toprule
    \textbf{Method} & \textbf{20ng} & \textbf{R8} & \textbf{R20} & \textbf{ohsumed} & \textbf{MR} & \textbf{Average}\\
    % \midrule
    % BERT (for reference) & 87.21 & 98.03 & 96.17 & 71.46 & 86.61 & 87.896\\
    \midrule
    WideMLP baseline & {\bf 83.31} & 97.27 & 93.89 & {\bf 63.95} & 76.72 & {\bf 83.03} \\
    WideMLP-b1.58-mean lr=$10^{-3}$ & 79.89 & {\bf 97.40} & 93.54 & 60.75 & {\bf 77.10} & 81.74 \\
    WideMLP-b1.58-median lr=$10^{-3}$ & 80.08 & 97.35 & {\bf 94.20} & 62.28 & 76.14 & 82.01 \\
    WideMLP-b1.58-median lr=$10^{-2}$ & 81.74 & 96.80 & 93.69 & 62.73 & 76.22 & 82.24 \\
    \bottomrule
    \end{tabular}
    \caption{Text Classification: A wide multi-layer perceptron on a bag-of-words (WideMLP) compared to corresponding BitNet variants WideMLP-b1.58-mean and WideMLP-b1.58-median. The results of the WideMLP baseline are taken from \citet{galke-scherp-2022-bag}.}\label{tab:textclf}
\end{table*}

\paragraph{Results}
Table~\ref{tab:textclf} shows the results. On average, WideMLP-b1.58-mean achieves 98.4\% of WideMLP's performance. WideMLP-b1.58-median achieves 98.8\% of WideMLP performance.
With a higher learning rate ($10^{-2}$), WideMLP-b1.58-median attains 99.0\% baseline performance. 

\subsection{BitNet in Graph Neural Networks}\label{sec:non-transformer:gnn}
We seek to understand how BitNet affects graph representation learning.
We evaluate 1.58-bit variants of simplified graph convolution on three commonly used citation network datasets to evaluate node classification in graphs: Cora, Citeseer, Pubmed under the split by \citet{pmlr-v48-yanga16}: using only the labels of 20 nodes per class for training.

\paragraph{Setup} As base graph neural network models we consider graph convolutional networks (GCN)~\citet{kipf2016semi} and simplified graph convolution (SGC)~\cite{sgc}. Specifically, we use a 2-layer ReLU-activated GCN and an SGC model with 2-hop neighborhood aggregation. In both models, we substitute the linear layers with $\operatorname{BitLinear}$ (2 for the GCN, and 1 for SGC). We then experiment with mean and median weight quantization measures. All models are trained for 100 epochs with learning rate 0.01. We report mean accuracy and 95\% confidence intervals across 10 repetitions.

\begin{table*}[ht]
    \centering
    \begin{tabular}{llllr}
    \toprule
     \textbf{Method} & \textbf{Cora} & \textbf{Citeseer} & \textbf{Pubmed} & \textbf{Avg.}  \\
     \midrule
        GCN baseline & $\mathbf{78.57} \pm 0.49$ & $63.76 \pm 0.48$ & $\mathbf{76.02} \pm 0.19$ & $\mathbf{72.78}$ \\
        GCN-b1.58-mean & $76.03 \pm 0.50$ & $\mathbf{65.83} \pm 0.45$ & $73.51 \pm 0.47$ & $71.79$ \\
        GCN-b1.58-median & $75.76 \pm 0.32$ & $65.60 \pm 0.65$ & $74.42 \pm 0.39$ & $71.93$ \\
        \midrule
     % k=2, lr = 0.01
        SGC baseline& $77.07 \pm 0.15$ & $\mathbf{63.66} \pm 0.18$ & $\mathbf{75.63} \pm 0.08$ & $\mathbf{72.12}$ \\
        SGC-b1.58-mean& $77.31 \pm 0.38$ & $59.31 \pm 1.82$ & $75.22 \pm 0.33$ & $70.61$ \\
        SGC-b1.58-median& $\mathbf{77.46} \pm 0.75$ & $61.31 \pm 1.90$ & $74.42 \pm 0.26$ & $71.06$ \\
        \bottomrule
        \end{tabular}
    \caption{
    Node classification. Graph convolutional networks (GCN) and Simplified Graph Convolution (SGC) compared to their BitNet variants (*-b1.58-mean and *-b1.58-median). Mean accuracy across 10 runs with 95\% confidence intervals estimated via 1.96 standard error of the mean.}
    \label{tab:nodeclf}
\end{table*}

\paragraph{Results}
Table~\ref{tab:nodeclf} shows the results. 
For both GCN and SGC, the b1.58-mean and b1.58-median variants yield accuracy scores very close to their full-precision counterparts, sometimes even higher SGC-b1.58-median on Cora and GCN-b.158-mean on Citeseer).
GCN-b1.58-median achieves 98.8\% relative performance to GCN and SGC-b1.58-median achieves 98.5\% of SGC's performance.

\subsection{Summary and Interim Discussion}
What can we learn from this set of experiments: In the toy \xor{} setting, we found that BitNet models need a higher number of parameters, or a higher learning rate as suggested by prior work on BitNet~\cite{wang2023bitnet}. However, in practical settings, such as text classification and node classification,  BitNet variants yield almost the same performance as their respective baselines. We hypothesize that this is because the models are sufficiently overparameterized. As for the difference between mean and median quantization schemes, we find that both options lead to similar performance. In the following, we will proceed with the median option only.

\section{BitNet in Transformer Architectures}\label{sec:transformer}
In this section, we present our comparisons training encoder, decoder, and encoder-decoder architectures employing BitLinear layers with the robust AbsMedian quantization for 1.58-bit weights and AbsMax quantization for 8-bit activations.

We will demonstrate that the performance of b1.58 in encoder-only architectures aligns with previous results on decoder-only architectures. 
In encoder-decoder architectures, we will show even superior performance over the 16-bit counterparts. Finally, we further explore this potential regularization effect of 1.58-bit training, which appears to prevent or at least delay overfitting.

% \rc{
% \begin{itemize}
%     \item Objectives to investigate
%     How does dropout correlate with regularization of using 1.58-bit weights.
%     \begin{itemize}
%         \item Dropout vs quantisation - does it work as a regularizer?
%         \item Influence in the forward pass of the dropout / Quantization
%       \item  Note: Peter used less dropout on MNIST experiments, we suspect 1.58 bits works as a regularizer.
%     \end{itemize}
%      \item  We have seen indications that larger networks are needed to obtain similar performance to full-precision training (we know that from reloaded)
%      \begin{itemize}
%          \item Does the parameters function / role change?
%          \item Do we need more parameters / higher capacity networks in general? (Vanilla Transformer!)
%      \end{itemize}
%      \item How does the activation range effect the performance? (Seems not to provide much $\rightarrow$ can we somehow visualize)
%      \item Can we visualize the difference in network "encoding" when using 1.58bit vs full resolution.(New explainable studies talk about the vector-space might not be 100\% perpendicular but rather within 89-91 degrees. This is suggested to be the key to the scalability of LLMs. We could address insights into the characteristics of 1.58bit and why it performs quite well. Further, this might help us to understand the limitations we try to understand in this work.)
% \end{itemize}
% }

\subsection{Encoder Models}\label{sec:transformers:encoders}
\paragraph{Setup}
We conduct experiments with the BERT-base encoder architecture demonstrating the scaling-law, already defined for decoder models by prior work \cite{nielsen2024bitnetb158reloadedstateoftheart}. We employ the Cramming framework \cite{geiping2023cramming} for experiments with increasing size of hidden dimensions for both 16-bit and 1.58-bit. Collectively, we use 12 attention heads and 16 encoder layers. We employ a learning rate of 0.001,  the \texttt{AdamW}-optimizer \cite{loshchilov2017decoupled} and a batch-size of 8192. We use the bert-o4 config\footnote{https://github.com/JonasGeiping/cramming/blob/main/ \newline cramming/config/train/bert-o4.yaml}. The dataset is tokenized using WordPiece \cite{wu2016googlesneuralmachinetranslation}.
The experiments are conducted with an internal dataset consisting of approximately 80\% Danish texts mixed with 20\% Norwegian, Swedish, German, and English over 6.2 million documents.

\begin{figure*}[ht]
    \centering
    \caption{Scaling Behavior of 16-bit and 1.58-bit (median) BERT Encoder Train Loss over $150$K train-steps. Smoothing applies a Savitzky-Golay filter with a window size of 1000 and a polynomial order of 2.}
    \begin{subfigure}[b]{0.45\textwidth}
        \centering
        \includegraphics[width=\textwidth]{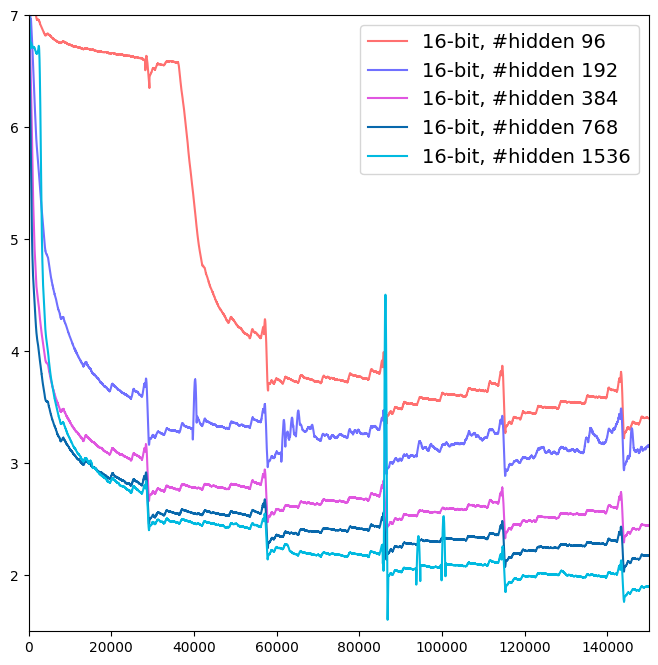}
        \caption{Scaling for 16 bit}
        \label{fig:encoder_scaling_16_bit}
    \end{subfigure}
    \hfill
    % Subfigure 2
    \begin{subfigure}[b]{0.45\textwidth}
        \centering
        \includegraphics[width=\textwidth]{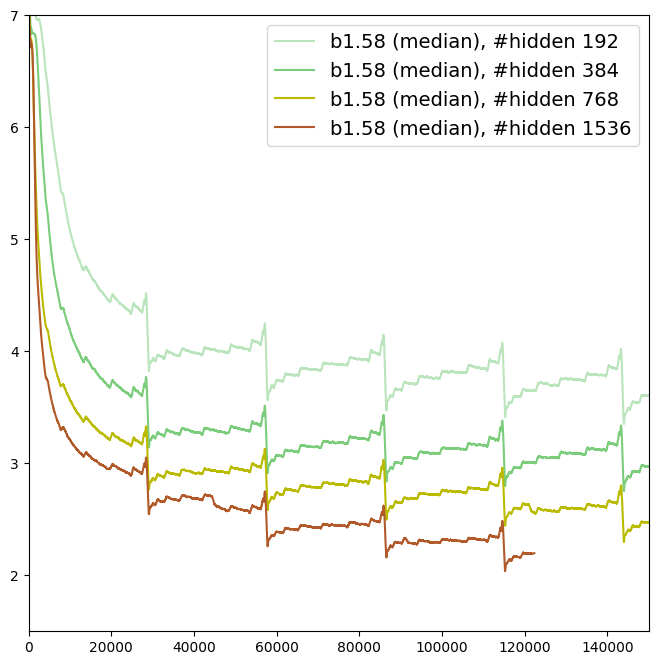} 
        \caption{Scaling for 1.58 bit (median)}
        \label{fig:encoder_scaling_158_bit}
    \end{subfigure}
    \label{fig:encoders_scaling}
\end{figure*}

\paragraph{Results}
We compare the 16 bit and b1.58 performances in Figures \ref{fig:encoder_scaling_16_bit} and \ref{fig:encoder_scaling_158_bit}. We observe that encoders scale as outlined by prior work and require  approximately double the hidden layer size to achieve performance comparable to 16-bit versions. We see a hidden layer size of 384 in b1.58 archives performance similar to hidden size of 192 in 16-bit. This trend is observed across other pairs of hidden sizes, e.g., 768 and 384 as well as 1536 and 768.

% \subsubsection{Encoder Partial Quantization} % encoder-only dense, full, attention, none quant
% How does partial b1.58 quantization effect Encoders? Shown in Figure \ref{fig:encoder_partial_quant}, we see that it does in fact not have much influence of either training-behavior or performance. \jacob{Is this interesting at all? Maybe in story about the o\_proj issues in T5}

% \begin{figure}
%     \centering
%     \includegraphics[width=0.95\linewidth]{plots/encoder_dense_full_attention_none_quant.png}
%     \caption{Partial Quantization of BERT Encoder}
%     \label{fig:encoder_partial_quant}
% \end{figure}

% We can confirm that no scaling law is needed.
% we are not training the fulll dataset so we are not seeing an effect on seeing samples / data twice.
% Data multiave exclusive, but from the same discription 
% /home/jps/nanoT5/.aim/
% Both validation and train loss for the paper
% paper for cite: validation is correlated wiht the donwstream

\subsection{Encoder-Decoder Models}\label{sec:transformer:endec}
\begin{figure*}[ht]
    \centering
    \caption{Scaling Behavior of 16-bit and b1.58 (median) employed in T5v1.1-Base Encoder-Decoder Architecture. 16-bit and b1.58 employed throughout the network over increasing hidden size. \texttt{d\_ff} denotes the hidden size of the FFN within each encoder and decoder stack.}
    \begin{subfigure}[b]{0.45\textwidth}
        \centering
        \includegraphics[width=\textwidth]{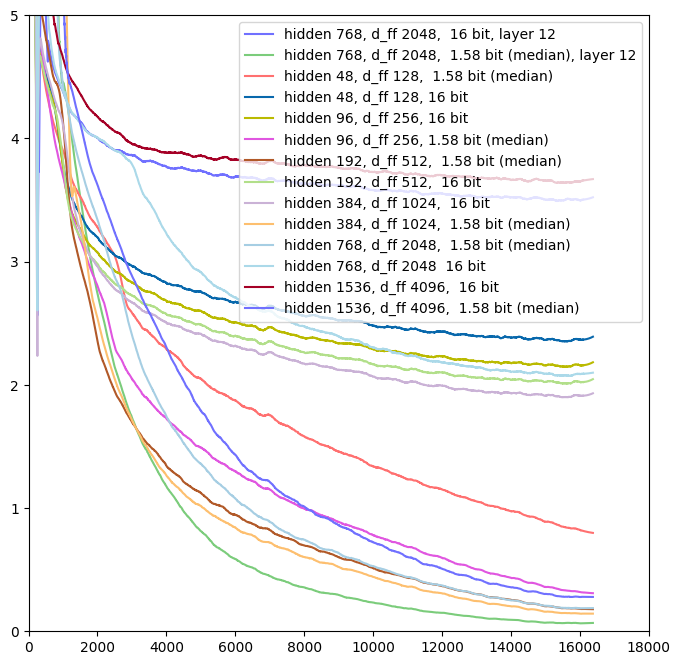}
        \caption{Training Loss. \texttt{[savitzky-golay, 512, 2]}}
        \label{fig:endec_train_loss_scaling}
    \end{subfigure}
    \hfill
    % Subfigure 2
    \begin{subfigure}[b]{0.45\textwidth}
        \centering
        \includegraphics[width=0.98\textwidth]{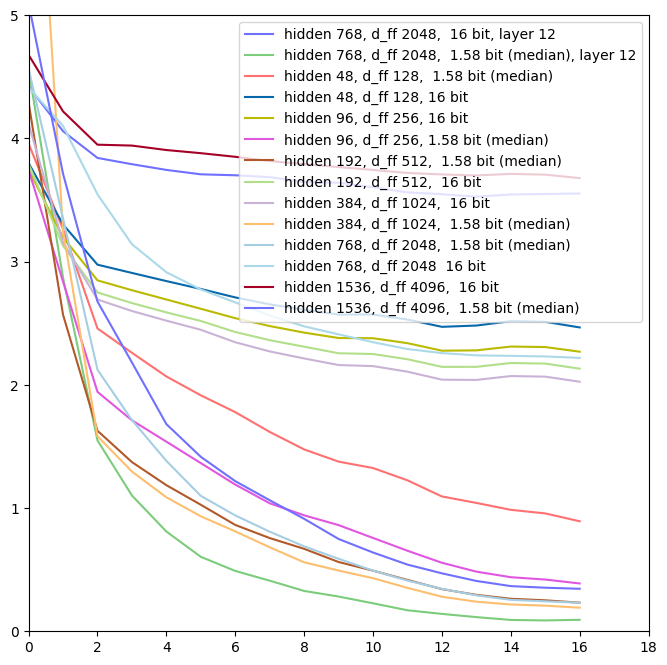} 
        \caption{Validation Loss. \texttt{[savitzky-golay, 4, 2]}}
        \label{fig:endec_eval_loss_scaling}
    \end{subfigure}
    \label{fig:endec_loss_scaling}
\end{figure*}

\paragraph{Setup}
We investigate the impact of 1.58-bit quantization on on encoder-decoder models. Specifically, we apply b1.58 to the task of masked language modelling, utilizing the NanoT5 framework \cite{nawrot-2023-nanot5} with a \texttt{T5v1\_1-base} architecture with the \texttt{mT5}  SentencePiece \cite{kudo2018sentencepiece} tokenizer, employing a masking-probability of 15\%. We used a learning rate of $0.02$, a batch size of $128$, and no weight decay, with a \texttt{cosine} learning rate scheduler and the \texttt{adamwscale} optimizer \cite{nawrot-2023-nanot5}. The model architecture includes 6 layers in both the encoder and decoder stacks (Fig. \ref{fig:endec_loss_scaling}). Hidden sizes are varied between 48 and 1536. Experiments are conducted on the dataset described in Section \ref{sec:transformers:encoders}.

% Hidden sizes are set to 768 and 1536, with a consistent feed-forward network hidden size of 4096. 
% For the module and layer-wise experiments we also employ 6 layers both for encoder and decoder. Hidden sizes and feed-forward network hidden size  with a consistent 128 and 512, respectively. 
% Here, we conducted scaling experiments using the same dataset as in Section \ref{sec:transformers:encoders}. To further investigate module- and layer-wise quantization, we employ the standard \texttt{WikiText2} dataset, consisting of over 100 million tokens \cite{merity2016pointer} when tokenized using SentencePiece \cite{kudo2018sentencepiece}.

\paragraph{Results}
We compare the 16 bit and 1.58 bit performance in Figure \ref{fig:endec_loss_scaling}. In Figure \ref{fig:endec_train_loss_scaling} we compare different hidden size's performance between 16-bit and 1.58-bit (median). We observe that 1.58-bit outperforms all 16-bit versions. Further, as expected, we see that 12 layers performs better than the default 6 layers. We refer too Figure \ref{fig:endec_eval_loss_scaling} evaluating the models on held-out validation set, however, coming from the same distribution. We observe that the validation loss aligns almost perfectly with their corresponding training loss curve, likely indicative of decent in-distribution generalization and decent downstream task performance \cite{pmlr-v202-liu23ao}.
%\todo{Did we have a citation that argued that this correlation, would yield promissing results on downstream tasks?} 

% We compare the 16 bit and 1.58 bit performance in Figures \ref{fig:endec_train_loss_scaling} and \ref{fig:endec_module_wise}. Observing Figure \ref{fig:endec_train_loss_scaling} we compare different hidden size's performance between 16-bit and b1.58. We observe 16-bit with hidden size 768 is superior, interestingly closely followed by hidden size 768 with b1.58 employed in dense layers. We observe the same tendency for hidden size 1536 albeit with a larger difference. Employing b1.58 (median) collectively in all dense layers evidently yields worse performance show-cased by hidden sizes 768 and 1536. 

% Investigation why employing b1.58 beyond the network's dense layers yields a drastic performance loss, we refer to Figure \ref{fig:endec_module_wise}, which visualizes experiments with different combinations of b1.58 employed throughout the network. \texttt{Dense}, \texttt{S} and \texttt{C} denote dense layers, self-attention, and cross-attention layers, respectively. \texttt{q}, \texttt{k}, \texttt{v} and \texttt{o} denotes query, key, value and output projection matrices utilized in self- and cross-attention mechanisms. We observe that employing b1.58 in any output-projection (\texttt{o}) yields a significant performance penalty. We further note, that employing the b1.58 in the cross-attention o-projection and not in the self-attention counter-part aids performance significantly, however with full 16-bit in that projection being clearly superior.

\subsection{1.58 bit is a Regularizer} \label{sec:regularizer}
\paragraph{Setup}
To better understand why the 1.58-bit language models even outperform the 16-bit models, we hypothesize that there could be a regularization effect of 1.58-bit quantization.
To further investigate this regularization effect, we run experiments using an Open Language Models (OLMo) model~\cite{OLMo} with 1B parameters. We employ b1.58 with weight-only quantization
%\todo{Jacob - are we sure about this? I was thinking I used 8-bit activation quantization, too. -> Hi, but if activation\_measure is None, we do not use the activation range? [{"activation\_measure": None, "activation\_range": 8, "match\_name": "", "weight\_measure": "AbsMedian", "weight\_range": 1.58}]}
using AbsMedian with a $1.58$ weight range. We train on the dataset described in Section \ref{sec:transformers:encoders} with OLMo's standard hyperparameters
%\footnote{https://github.com/allenai/OLMo/blob/main/configs/official/OLMo-1B.yaml}
. We train all models with a batchsize of 2048.
% \todo[color=red]{FIXME: Danish dataset - fixed.}.

\begin{figure*}[ht]
    \centering
    \caption{Regularization Effect of b1.58 (median), OLMo 1B model trained on the internal dataset described in Section \ref{sec:transformers:encoders}. Smoothing is applied using a Savitzky-Golay filter with a window size of 1000 and polynomial order 2.}
    \begin{subfigure}[b]{0.45\textwidth}
        \centering
        \includegraphics[width=\textwidth]{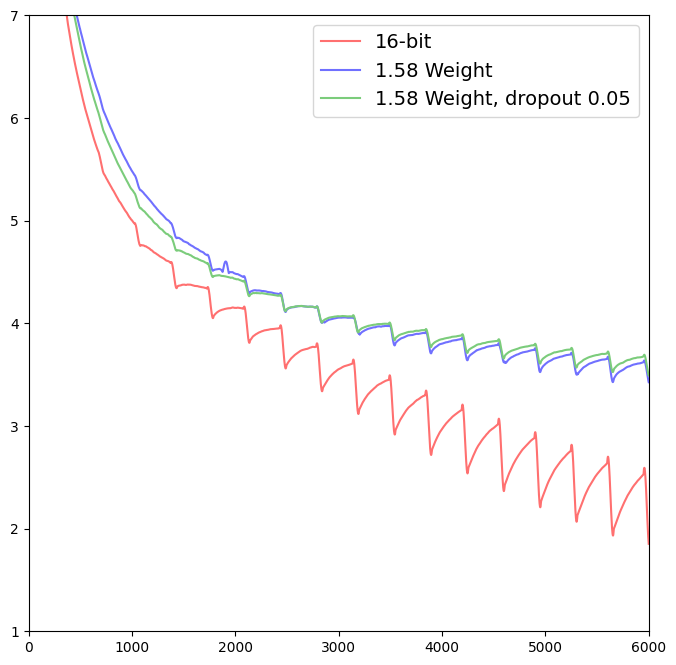}
        \caption{Training Loss}
        \label{fig:decoder_regulization_training}
    \end{subfigure}
    \hfill
    % Subfigure 2
    \begin{subfigure}[b]{0.45\textwidth}
        \centering
        \includegraphics[width=\textwidth]{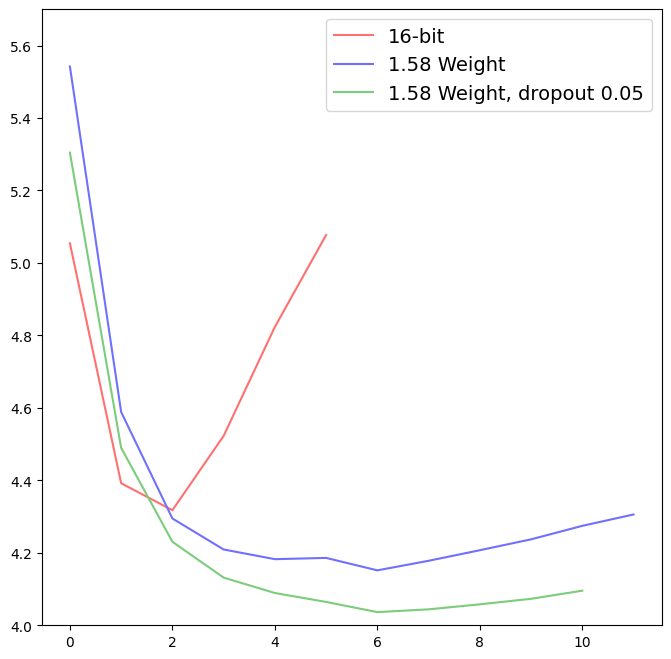} 
        \caption{Validation Loss}
        \label{fig:decoder_regulization_evaluation}
    \end{subfigure}
    \label{fig:decoder_regularization}
\end{figure*}

% context length: 2048
% 16 layers
% n_heads 16
\paragraph{Results}
In Figure \ref{fig:decoder_regularization}, we evaluate the abilities of the OLMo models to fit the training and validation datasets. In Figure \ref{fig:decoder_regulization_training}, as expected, we observe that 16-bit performs best, with b1.58 performing robustly with only negligible performance loss, even when a dropout of $0.05$ is applied. 

However, in Figure \ref{fig:decoder_regulization_evaluation}, we see that b1.58 delays overfitting to the training dataset. We attribute this to a regularization effect the quantization must introduce, making the b1.58 versions superior on the evaluation set. Note that dropout as another form of regularization likewise aids the performance on the evaluation set for both 16-bit and b1.58 models.

% \begin{itemize}
%     \item Plots:
%     \begin{itemize}
%         \item Scaling 16 bits
%         \item Scaling law? $\rightarrow$ does not make much sense
%         \item Showcase working and non-working quantizations of sub networks. 
%         \item Layerwise Quantization of T5
%     \end{itemize}
% \end{itemize}

% Dmodel
% FF dim

% \rc{
% \begin{itemize}
%     \item Dense-layers 
%     \item Bitlinearize in selected Encoder/ Decoder Layers
%     \begin{itemize}
%         \item E.g. 5 bitlinearized succeeded by one in full resolution (or the other way around). What works?
%     \end{itemize}
%     \item O-projection in self-attention.
%     \begin{itemize}
%         \item Can we loose this output projection? What does it mean for performance. (it is not "native" to the Transformer and is T5 design)
%     \end{itemize}
% \end{itemize}
% }

\section{Discussion}\label{sec:discussion}
We have shown that BitNet models can be trained to yield commensurate accuracy with their standard 16/32-bit counterparts. In T5-like architectures, the BitNet models perform even better than the 16-bit models -- which we attribute to a regularization effect. These results highlight that 1.58-bit quantization aware training methods could greatly reduce memory requirements and increase throughput at inference time for a wide range of models. 
%\todo{lukas stopped here}

% \paragraph{BitNet models trained from scratch are on par with their half \rc{or full-precision} counterparts}
% \paragraph{BitNet models need higher capacity to compensate for reduced precision}
Throughout Sections \ref{sec:non-transformer} and \ref{sec:transformer} we have shown that 1.58-bit models yield performance comparable to standard 16/32-bit models -- both for non-transformer and transformer models. In some cases, the 1.58-bit models even outperform 16-bit models: on multilayer perceptrons (see Section~\ref{sec:non-transformer:mlp}), graph neural networks (Section~\ref{sec:non-transformer:gnn}, and encoder-decoder language models (Section~\ref{sec:transformer:endec}).

% lg: too fine-grained here? 
% As reported in Section \ref{sec:x-or}, we demonstrate perfect performance in solving the X-OR task using AbsMean quantization, albeit a lower performance using AbsMedian, which we show can be compensated by introducing capacity through a doubling of the hidden size from 8 to 16. However, this latter setting yields an unstable training trajectory, which can be improved further by increasing hidden size to 32, concluding at 100\% accuracy. \todo{Stil valid? lukas: yes but also higher lr does the trick with 8 hidden units}
Comparing the two weight quantization schemes  for BitNet: AbsMedian and AbsMean, we have observed that AbsMedian is on par with AbsMean.
Prior work has shown that AbsMedian performs better in some situations \cite{nielsen2024bitnetb158reloadedstateoftheart}, conjecturing AbsMedian to be more resilient to weight-updates, which allows for higher variance without noticeable effect on the scaling factor. While this may be a factor for models with millions of parameters, the miniature models involved in solving the \xor{} task seem to be extraordinarily sensitive when the scaling factor is computed on a much smaller sample. We conjecture that this is the source of the instability observed in some configurations (e.g, hidden size 8 and 16 with low learning rate), which we have observed to be dampened when increasing the hidden size or the learning rate.
In the more practical scenarios, we see that AbsMean and AbsMedian quantization are close to each other with a $0.6\%$ drop in accuracy for MLPs and within each other's confidence intervals for GNNs.
This is in line with prior work for small language and vision models \cite{nielsen2024bitnetb158reloadedstateoftheart} and underlines the strong potential of b1.58 models. 

In Section \ref{sec:transformers:encoders}, we conducted experiments investigating how encoder-only transformers scale with b1.58. As stated by prior work, while b1.58 still relies on 16-bit weights as shadow weights quantized to ternary during forward passes, the capacity of each weight in linear layers is reduced, potentially affecting the performance of the overall network \cite{nielsen2024bitnetb158reloadedstateoftheart}. In line with such results, we also observe the need to re-introduce capacity in some cases. This is demonstrated for encoders in Figures \ref{fig:encoder_scaling_16_bit} and \ref{fig:encoder_scaling_158_bit}, where we report the scaling for 16-bit and b1.58 with AbsMedian over increasing hidden sizes. We see the need to utilize a hidden size of 192 in 1.58-median to gain the same performance as 16-bit with a hidden size of 96. This also holds for hidden layer sizes of 384, 768 and 1536 for 1.58-median paired with 192, 368 and 768 for 16-bit, respectively. This confirms that there is a similar scaling law for encoder-only transformers as for decoder-only transformers. In general, we believe that whether to increasing model capacity depend less on the particular transformer architecture but rather on the complexity of the downstream task and the given model size. As shown by prior work, LLMs do not utilize all parameters effectively and comprise redundant layers~\cite{ashkboos2024slicegpt,he2024matterstransformersattentionneeded}, prompting us to believe these existing parameters can be utilized as active capacity when 1.58-bit is employed, explaining the reduction in performance gap when LLM size increases. % these considerations is in \cite{nielsen2024bitnetb158reloadedstateoftheart} 
We do not believe that the potential need for additional layers to be an obstacle for b1.58 models as they can be deployed in a lot more efficient manner using dedicated kernels, where standard 16/32-bit matrix multiplication might be replaced with signed addition for increased hardware utilization.

In Section \ref{sec:transformer:endec}, we report that findings for the encoder-decoder transformer architectures do not directly align with the findings for the separate encoder-only and decoder-only architectures. Demonstrated in Figure \ref{fig:endec_eval_loss_scaling}, employing 1.58-bit throughout the model generally enhance performance significantly, validated across both the training and validation sets, where we observe a very strong correlation. The plots does not include a full epoch worth of steps, therefore, we cannot contribute this performance to any affect of seeing the samples multiple times. Together, these findings suggests, that the scaling law for encoder and decoder models is not applicable for encoder-decoder models as the architecture. Not only, does not need an increased hidden size, but also outperforms the 16-bit counterpart. We encourage future work to investigate why 1.58-bit aid the training.

In Figure \ref{fig:endec_train_loss_scaling}, we see ``knees'' forming in multiple loss graphs, where the training exhibits a sudden drop in loss at a certain point of time e.g. in the light teal and brown graph. Depending on the network size and quantization, we see this time shifting in the first 4000 steps. This phenomenon has been observed in prior work \cite{chen2024sudden} on MLMs, describing the learning-phase before the drop as the syntactic attention structure (SAS) phase, where the networks learns the syntactic structure and relations between components in the data before starting to focus on the task, yielding the steep drop in loss. We believe that these knees constitutes the end of the SAS-phase and the start of the ``capabilities onset'' phase. The encoder needs to learn the syntactic structure and be able to encode the input to a latent vector with high enough semantic value for the decoder to learn something meaningful, yielding a lower loss. The decoder is highly dependent on the encoder to deliver such a representation fairly early for the learning to be efficient, and we believe this mechanism, together with the identical task, is the reason why we see similar behaviour as in \cite{chen2024sudden}.
%\todo{Should we still consider syntax learning?}

% Similar to described results, employing quantization throughout the first four layers and further in the last-two unless the \texttt{o}-projection works yields a performance close the 16-bit version. This underlines our findings on the o-projection, making the most quantized network competitive to 16-bit.

% \paragraph{Down projection layers are difficult to bitlinearize}
% \rc{Should we comment on this bottle-neck layer might be troublesome due to lost capacity in preceding modules?}
% \todo{do we understand this enough to make an educated guess?}

% \paragraph{b158 as a regularizer}
% \rc{
%     \begin{itemize}
%         \item Why the O-layer? - Is that the bottleneck or are we "moving" the issue onto that layer when altering capacity in the rest of the layer.
%         \item We might get an insight on these by bitlineraize layers selectively. e.g. trying preceding layer and succeeding layers separately. 
%     \end{itemize}
% }
% \todo{write a short summary}
% \rc{Check dataset again}
In Section \ref{sec:regularizer}, we report on the observed regularization effect of b1.58 when training a 1B parameter OLMo model on the internal dataset, described in Section \ref{sec:transformers:encoders}.  We observe that 16-bit fit the training set both better and faster but also declines on the evaluation set, i.e., it is overfitting. The quantized models instead overfit less/later, which supports the idea of quantization acting as a regularizer.

We see that 1.58-bit without and with dropout perform better and best, respectively, on the validation set. These results indicate that the coarseness of the ternary quantization also contributes positively to model regularization (see Figure \ref{fig:decoder_regularization}), which is beneficial for good generalization performance. 
\section{Conclusion}
We conducted a bottom-up investigation of 1.58-bit quantization-aware training for a range of both non-transformer and transformer models, demonstrating very competitive performance for multi-layer perceptrons, graph neural networks, encoder-only, and encoder-decoder architectures when compared to 16 and 32-bit. 
Specifically, we find that b1.58-bit training works well beyond language models, demonstrating competitive performance in bag-of-words MLPs for text classification and graph neural networks for node classification. Median quantization seems on par or better than mean in most real-world scenarios. We further analyze the ``BitNet Scaling Law'' for encoder-only models, showing that 1.58-bit models match the performance of standard precision models when the hidden size is twice as large, aligning with similar observations for decoder-only models. For encoder-decoder models, we find that that no such scaling law is applicable, as b1.58 consistently performs better than 16-bit. We encourage future research to investigate the mechanisms that allow b1.58 with its ternary weights improve performance significantly within the encoder-decoder architecture.  
Finally, there seems to be a regularization effects of 1.58-bit quantization-aware training that helps generalization. Yet, more research is needed to further investigate this regularization effect.

\section{Limitations}
To fully harvest the potential of 1.58-bit quantization regarding memory use, latency, and throughput, there is a need for specialized kernels. Lastly, in our experiments on language models, we employ the validation loss as proxy for downstream performance. While this is common practice in language modeling, there could be effects of quantization harming the correlation of validation loss and downstream performance.
%\todo{Limitations and Ethical Considerations don't count against page limit}

\section{Ethical Considerations}
Quantization-aware training for b1.58 requires a slight increase in training
resources due to having to quantize and scale weights and activations. However,
in exchange, required resources at inference time are greatly reduced. Thus,
with specialized kernels, b1.58 models would contribute to more sustainable
inference and serving of large models.
Reducing the memory footprint also allows researchers and practitioners to carry out more computation locally and therefore may help alleviate privacy concerns.

\bibliographystyle{acl_natbib}
\bibliography{sample.bib}

\begin{thebibliography}{27}
\expandafter\ifx\csname natexlab\endcsname\relax\def\natexlab#1{#1}\fi

\bibitem[{Ashkboos et~al.(2024)Ashkboos, Croci, do~Nascimento, Hoefler, and Hensman}]{ashkboos2024slicegpt}
Saleh Ashkboos, Maximilian~L. Croci, Marcelo~Gennari do~Nascimento, Torsten Hoefler, and James Hensman. 2024.
\newblock \href {http://arxiv.org/abs/2401.15024} {Slicegpt: Compress large language models by deleting rows and columns}.

\bibitem[{Ba et~al.(2016)Ba, Kiros, and Hinton}]{ba2016layer}
Jimmy~Lei Ba, Jamie~Ryan Kiros, and Geoffrey~E Hinton. 2016.
\newblock Layer normalization.
\newblock \emph{arXiv preprint arXiv:1607.06450}.

\bibitem[{Bal et~al.(2024)Bal, Jiang, and Sengupta}]{bal2024exploring}
Malyaban Bal, Yi~Jiang, and Abhronil Sengupta. 2024.
\newblock Exploring extreme quantization in spiking language models.
\newblock \emph{arXiv preprint arXiv:2405.02543}.

\bibitem[{Bengio et~al.(2013)Bengio, L{\'{e}}onard, and Courville}]{DBLP:journals/corr/BengioLC13}
Yoshua Bengio, Nicholas L{\'{e}}onard, and Aaron~C. Courville. 2013.
\newblock \href {http://arxiv.org/abs/1308.3432} {Estimating or propagating gradients through stochastic neurons for conditional computation}.
\newblock \emph{CoRR}, abs/1308.3432.

\bibitem[{Chen et~al.(2024)Chen, Shwartz-Ziv, Cho, Leavitt, and Saphra}]{chen2024sudden}
Angelica Chen, Ravid Shwartz-Ziv, Kyunghyun Cho, Matthew~L Leavitt, and Naomi Saphra. 2024.
\newblock \href {https://openreview.net/forum?id=MO5PiKHELW} {Sudden drops in the loss: Syntax acquisition, phase transitions, and simplicity bias in {MLM}s}.
\newblock In \emph{The Twelfth International Conference on Learning Representations}.

\bibitem[{Frantar et~al.(2023)Frantar, Ashkboos, Hoefler, and Alistarh}]{frantar2023gptq}
Elias Frantar, Saleh Ashkboos, Torsten Hoefler, and Dan Alistarh. 2023.
\newblock \href {http://arxiv.org/abs/2210.17323} {Gptq: Accurate post-training quantization for generative pre-trained transformers}.

\bibitem[{Galke and Scherp(2022)}]{galke-scherp-2022-bag}
Lukas Galke and Ansgar Scherp. 2022.
\newblock \href {https://doi.org/10.18653/v1/2022.acl-long.279} {Bag-of-words vs. graph vs. sequence in text classification: Questioning the necessity of text-graphs and the surprising strength of a wide {MLP}}.
\newblock In \emph{Proceedings of the 60th Annual Meeting of the Association for Computational Linguistics (Volume 1: Long Papers)}, pages 4038--4051, Dublin, Ireland. Association for Computational Linguistics.

\bibitem[{Geiping and Goldstein(2023)}]{geiping2023cramming}
Jonas Geiping and Tom Goldstein. 2023.
\newblock Cramming: Training a language model on a single gpu in one day.
\newblock In \emph{International Conference on Machine Learning}, pages 11117--11143. PMLR.

\bibitem[{Groeneveld et~al.(2024)Groeneveld, Beltagy, Walsh, Bhagia, Kinney, Tafjord, Jha, Ivison, Magnusson, Wang, Arora, Atkinson, Authur, Chandu, Cohan, Dumas, Elazar, Gu, Hessel, Khot, Merrill, Morrison, Muennighoff, Naik, Nam, Peters, Pyatkin, Ravichander, Schwenk, Shah, Smith, Strubell, Subramani, Wortsman, Dasigi, Lambert, Richardson, Zettlemoyer, Dodge, Lo, Soldaini, Smith, and Hajishirzi}]{OLMo}
Dirk Groeneveld, Iz~Beltagy, Pete Walsh, Akshita Bhagia, Rodney Kinney, Oyvind Tafjord, A.~Jha, Hamish Ivison, Ian Magnusson, Yizhong Wang, Shane Arora, David Atkinson, Russell Authur, Khyathi~Raghavi Chandu, Arman Cohan, Jennifer Dumas, Yanai Elazar, Yuling Gu, Jack Hessel, Tushar Khot, William Merrill, Jacob~Daniel Morrison, Niklas Muennighoff, Aakanksha Naik, Crystal Nam, Matthew~E. Peters, Valentina Pyatkin, Abhilasha Ravichander, Dustin Schwenk, Saurabh Shah, Will Smith, Emma Strubell, Nishant Subramani, Mitchell Wortsman, Pradeep Dasigi, Nathan Lambert, Kyle Richardson, Luke Zettlemoyer, Jesse Dodge, Kyle Lo, Luca Soldaini, Noah~A. Smith, and Hanna Hajishirzi. 2024.
\newblock \href {https://api.semanticscholar.org/CorpusID:267365485} {Olmo: Accelerating the science of language models}.
\newblock \emph{arXiv preprint}.

\bibitem[{He et~al.(2024)He, Sun, Shen, and Li}]{he2024matterstransformersattentionneeded}
Shwai He, Guoheng Sun, Zheyu Shen, and Ang Li. 2024.
\newblock \href {http://arxiv.org/abs/2406.15786} {What matters in transformers? not all attention is needed}.

\bibitem[{Kingma and Ba(2015)}]{kingma2014adam}
Diederik~P Kingma and Jimmy Ba. 2015.
\newblock {Adam: A method for stochastic optimization}.
\newblock In \emph{Proceedings of the International Conference on Learning Representations}.

\bibitem[{Kipf and Welling(2016)}]{kipf2016semi}
Thomas~N Kipf and Max Welling. 2016.
\newblock Semi-supervised classification with graph convolutional networks.
\newblock \emph{arXiv preprint arXiv:1609.02907}.

\bibitem[{Kudo(2018)}]{kudo2018sentencepiece}
T~Kudo. 2018.
\newblock Sentencepiece: A simple and language independent subword tokenizer and detokenizer for neural text processing.
\newblock \emph{arXiv preprint arXiv:1808.06226}.

\bibitem[{Li and Gu(2023)}]{li2023vit}
Zhikai Li and Qingyi Gu. 2023.
\newblock I-vit: integer-only quantization for efficient vision transformer inference.
\newblock In \emph{Proceedings of the IEEE/CVF International Conference on Computer Vision}, pages 17065--17075.

\bibitem[{Lin et~al.(2024)Lin, Tang, Tang, Yang, Chen, Wang, Xiao, Dang, Gan, and Han}]{lin2024awq}
Ji~Lin, Jiaming Tang, Haotian Tang, Shang Yang, Wei-Ming Chen, Wei-Chen Wang, Guangxuan Xiao, Xingyu Dang, Chuang Gan, and Song Han. 2024.
\newblock \href {http://arxiv.org/abs/2306.00978} {Awq: Activation-aware weight quantization for llm compression and acceleration}.

\bibitem[{Liu et~al.(2023{\natexlab{a}})Liu, Xie, Li, and Ma}]{pmlr-v202-liu23ao}
Hong Liu, Sang~Michael Xie, Zhiyuan Li, and Tengyu Ma. 2023{\natexlab{a}}.
\newblock \href {https://proceedings.mlr.press/v202/liu23ao.html} {Same pre-training loss, better downstream: Implicit bias matters for language models}.
\newblock In \emph{Proceedings of the 40th International Conference on Machine Learning}, volume 202 of \emph{Proceedings of Machine Learning Research}, pages 22188--22214. PMLR.

\bibitem[{Liu et~al.(2023{\natexlab{b}})Liu, Oguz, Zhao, Chang, Stock, Mehdad, Shi, Krishnamoorthi, and Chandra}]{liu2023llmqat}
Zechun Liu, Barlas Oguz, Changsheng Zhao, Ernie Chang, Pierre Stock, Yashar Mehdad, Yangyang Shi, Raghuraman Krishnamoorthi, and Vikas Chandra. 2023{\natexlab{b}}.
\newblock \href {http://arxiv.org/abs/2305.17888} {Llm-qat: Data-free quantization aware training for large language models}.

\bibitem[{Loshchilov(2017)}]{loshchilov2017decoupled}
I~Loshchilov. 2017.
\newblock Decoupled weight decay regularization.
\newblock \emph{arXiv preprint arXiv:1711.05101}.

\bibitem[{Ma et~al.(2024)Ma, Wang, Ma, Wang, Wang, Huang, Dong, Wang, Xue, and Wei}]{ma2024era}
Shuming Ma, Hongyu Wang, Lingxiao Ma, Lei Wang, Wenhui Wang, Shaohan Huang, Li~Dong, Ruiping Wang, Jilong Xue, and Furu Wei. 2024.
\newblock \href {http://arxiv.org/abs/2402.17764} {The era of 1-bit llms: All large language models are in 1.58 bits}.

\bibitem[{Nawrot(2023)}]{nawrot-2023-nanot5}
Piotr Nawrot. 2023.
\newblock \href {https://doi.org/10.18653/v1/2023.nlposs-1.11} {nano{T}5: Fast {\&} simple pre-training and fine-tuning of {T}5 models with limited resources}.
\newblock In \emph{Proceedings of the 3rd Workshop for Natural Language Processing Open Source Software (NLP-OSS 2023)}. Association for Computational Linguistics.

\bibitem[{Nielsen and Schneider-Kamp(2024)}]{nielsen2024bitnetb158reloadedstateoftheart}
Jacob Nielsen and Peter Schneider-Kamp. 2024.
\newblock Bitnet b1. 58 reloaded: State-of-the-art performance also on smaller networks.
\newblock In \emph{International Conference on Deep Learning Theory and Applications}, pages 301--315. Springer.

\bibitem[{Sundaram and Iyer(2024)}]{sundaram2024llavaolmobitnet1b}
Jainaveen Sundaram and Ravishankar Iyer. 2024.
\newblock Llavaolmobitnet1b: Ternary llm goes multimodal!
\newblock \emph{arXiv preprint arXiv:2408.13402}.

\bibitem[{Wang et~al.(2023)Wang, Ma, Dong, Huang, Wang, Ma, Yang, Wang, Wu, and Wei}]{wang2023bitnet}
Hongyu Wang, Shuming Ma, Li~Dong, Shaohan Huang, Huaijie Wang, Lingxiao Ma, Fan Yang, Ruiping Wang, Yi~Wu, and Furu Wei. 2023.
\newblock \href {http://arxiv.org/abs/2310.11453} {Bitnet: Scaling 1-bit transformers for large language models}.

\bibitem[{Wu et~al.(2019)Wu, Souza, Zhang, Fifty, Yu, and Weinberger}]{sgc}
Felix Wu, Amauri Souza, Tianyi Zhang, Christopher Fifty, Tao Yu, and Kilian Weinberger. 2019.
\newblock \href {https://proceedings.mlr.press/v97/wu19e.html} {Simplifying graph convolutional networks}.
\newblock In \emph{Proceedings of the 36th International Conference on Machine Learning}, volume~97 of \emph{Proceedings of Machine Learning Research}, pages 6861--6871. PMLR.

\bibitem[{Wu et~al.(2016)Wu, Schuster, Chen, Le, Norouzi, Macherey, Krikun, Cao, Gao, Macherey, Klingner, Shah, Johnson, Liu, Łukasz Kaiser, Gouws, Kato, Kudo, Kazawa, Stevens, Kurian, Patil, Wang, Young, Smith, Riesa, Rudnick, Vinyals, Corrado, Hughes, and Dean}]{wu2016googlesneuralmachinetranslation}
Yonghui Wu, Mike Schuster, Zhifeng Chen, Quoc~V. Le, Mohammad Norouzi, Wolfgang Macherey, Maxim Krikun, Yuan Cao, Qin Gao, Klaus Macherey, Jeff Klingner, Apurva Shah, Melvin Johnson, Xiaobing Liu, Łukasz Kaiser, Stephan Gouws, Yoshikiyo Kato, Taku Kudo, Hideto Kazawa, Keith Stevens, George Kurian, Nishant Patil, Wei Wang, Cliff Young, Jason Smith, Jason Riesa, Alex Rudnick, Oriol Vinyals, Greg Corrado, Macduff Hughes, and Jeffrey Dean. 2016.
\newblock \href {http://arxiv.org/abs/1609.08144} {Google's neural machine translation system: Bridging the gap between human and machine translation}.

\bibitem[{Xu et~al.(2023)Xu, Xie, Gu, Chen, Chang, Zhang, Chen, Zhang, and Tian}]{xu2023qaloraquantizationawarelowrankadaptation}
Yuhui Xu, Lingxi Xie, Xiaotao Gu, Xin Chen, Heng Chang, Hengheng Zhang, Zhengsu Chen, Xiaopeng Zhang, and Qi~Tian. 2023.
\newblock \href {http://arxiv.org/abs/2309.14717} {Qa-lora: Quantization-aware low-rank adaptation of large language models}.

\bibitem[{Yang et~al.(2016)Yang, Cohen, and Salakhudinov}]{pmlr-v48-yanga16}
Zhilin Yang, William Cohen, and Ruslan Salakhudinov. 2016.
\newblock \href {https://proceedings.mlr.press/v48/yanga16.html} {Revisiting semi-supervised learning with graph embeddings}.
\newblock In \emph{Proceedings of The 33rd International Conference on Machine Learning}, volume~48 of \emph{Proceedings of Machine Learning Research}, pages 40--48, New York, New York, USA. PMLR.

\end{thebibliography}





% LNCS Homepage, \url{http://www.springer.com/lncs}, last accessed 2023/10/25
% \end{thebibliography}
\end{document}